\pdfoutput=1

\documentclass[11pt]{article}

\usepackage{emnlp2021}

\usepackage{times}
\usepackage{booktabs}
\usepackage{latexsym}
\usepackage[T1]{fontenc}

\usepackage[utf8]{inputenc}

\usepackage{microtype}
\usepackage{graphicx}

\usepackage{amsmath}
\usepackage{ amssymb }
\usepackage{wrapfig}
\usepackage{hyperref}
\usepackage[section]{placeins}
\usepackage{algorithmic}
\usepackage{algorithm}

\newcommand{\E}{\mathbb{E}}

\definecolor{aqua}{rgb}{0.0, 1.0, 1.0}

\usepackage[textsize=large]{todonotes}
\renewcommand{\vec}[1]{{\boldsymbol{\mathbf{#1}}}}

\newcommand{\Loss}{\mathcal{L}}
\newcommand{\Reward}{\mathcal{R}}

\newcommand\blfootnote[1]{%
  \begingroup
  \renewcommand\thefootnote{}\footnote{#1}%
  \addtocounter{footnote}{-1}%
  \endgroup
}
\title{Doubly-Trained Adversarial Data Augmentation\\ for Neural Machine Translation}

\author{Weiting Tan\quad Shuoyang Ding\quad Huda Khayrallah$^*$ \quad Philipp Koehn\\
  Center for Language and Speech Processing, Johns Hopkins University \\
  \texttt{\{wtan12, dings, huda, phi\}@jhu.edu}
}

\begin{document}
\maketitle

\begin{abstract}
Neural Machine Translation (NMT) models are known to suffer from noisy inputs. To make models robust, we generate adversarial augmentation samples that attack the model and preserve the source-side semantic meaning at the same time. To generate such samples, we propose a doubly-trained architecture that pairs two NMT models of opposite translation directions with a joint loss function, which combines the target-side attack and the source-side semantic similarity constraint. The results from our experiments across three different language pairs and two evaluation metrics show that these adversarial samples improve the model robustness. 
\end{abstract}

\section{Introduction\blfootnote{$^*$ Huda  Khayrallah is now at Microsoft.}}
When NMT models are trained on clean parallel data, they are not exposed to much noise, resulting in poor robustness when translating noisy input texts. Various adversarial attack methods have been explored for computer vision \cite{yuan2018adversarial}, including Fast Gradient Sign Methods \cite{goodfellow2015explaining}, and  generative adversarial networks  \cite[GAN;][]{goodfellow2014generative}, among others. Most of these methods are white-box attacks where model parameters are accessible during the attack so that the attack is much more effective. Good adversarial samples could also enhance model robustness by introducing perturbation as data augmentation \cite{goodfellow2014generative,chen2020realistic}. 

Due to the discrete nature of natural languages, most of the early-stage adversarial attacks on NMT focused on black-box attacks (attacks without access to model parameters) and use techniques such as string modification based on edit distance \cite{karpukhin2019training} or random changes of words in input sentence \cite{ebrahimi2018adversarial}). Such black-box methods can improve model robustness. However, model parameters are not accessible in black-box attacks and therefore restrict black-box methods' capacity. There is also work that tried to incorporate gradient-based adversarial techniques into natural languages processing, such as virtual training algorithm \cite{miyato2017adversarial} and adversarial regularization \cite{sato-etal-2019-effective}. These gradient-based adversarial approaches, to some extent, improve model performance and robustness. \citet{cheng2019robust,cheng2020advaug} further constrained the direction of perturbation with source-side semantic similarity and observed better performance.

Our work improves the gradient-based generation mechanism with a doubly-trained system, inspired by dual learning \cite{xia2016dual}. The doubly-trained system consists of a forward (translate from source language to target language) and a backward (translate target language to source language) model. After pretraining both forward and backward models, our augmentation process has three steps:
\begin{enumerate}
    \item \textit{Attack Step}: Train forward and backward models at the same time to update the shared embedding of source language (embedding of the forward model's encoder and the backward model's decoder).
    \item \textit{Perturbation Step}: Generate adversarial sequences by modifying source input sentences with random deletion and nearest neighbor search.
    \item \textit{Augmentation Training Step}: Train the forward model on the adversarial data.
\end{enumerate}

\noindent We applied our method on test data with synthetic noise and compared it against different baseline models. Experiments across three languages showed consistent improvement of model robustness using our algorithm.\footnote{code: https://github.com/steventan0110/NMTModelAttack}


\section{Background}
\paragraph{Neural Machine Translation}
Given a parallel corpus $S = {(\vec{x}^{(s)}, \vec{y}^{(s)})}_{s=1}^{|S|}$, for each pair of sentences, the model will compute the translation probability: $$ P(\vec{y}|\vec{x};\theta) = \prod_{n=1}^{N} P(\vec{y}_n | \vec{x}, \vec{y}_{<n};\theta)$$ where $\theta$ is a set of model parameters and $\vec{y}_{<n} = \vec{y}_1, \cdots, \vec{y}_n$ is a partial translation until $n^{th}$ position. The training objective is to maximize the log-likelihood of $S$, or equivalently, minimizing the negative log-likelihood loss (NLL): 

\begin{equation}
    \begin{aligned}
        \Loss_{\text{NLL}}(\theta) &= -\sum_{s=1}^{S}\log P(\vec{y}^{(s)} | \vec{x}^{(s)};\theta) \\
        &= -\sum_{s=1}^{S} \sum_{n=1}^{N(s)} \sum_{e}^{V} y^{(s)}_{n, e}\log  \hat{y}^{(s)}_{n, e} \\
        \hat{\theta}_{\text{NLL}} &= \underset{\theta}{\text{argmin}}  \{\Loss_{\text{NLL}}(\theta)\} 
    \end{aligned}
\end{equation}

Where $N(s)$ is length of $y^{(s)}$, $V$ is the vocabulary, $y^{(s)}_{n, e}=1$ if the $n^{th}$ word in $y^{(s)}$ is $e$ and $\hat{y}^{(s)}_{n, e}$ is the predicted probability of word entry e as the $n^{th}$ word for $y^{(s)}$.
To train the NMT model, we compute the partial of the loss over model parameter $\frac{\partial \Loss_{\text{NLL}}(\theta)}{\partial \theta}$ to update $\theta$. Note that sometimes we use NLL and cross entropy interchangeably because the last layer of our model is always softmax.

\paragraph{Minimum Risk Training (MRT)} \citet{shen-etal-2016-minimum} introduces evaluation metric into loss function and assume that the optimal set of model parameters will minimize the expected loss on the training data. The loss function is defined as $\Delta (\vec{y} , \vec{y}^{(s)})$ to measure the discrepancy between model output y and gold standard translation $\vec{y}^{(s)}$. It can be any negative sentence-level evaluation metric such as BLEU, METEOR, COMET, BERTScore, \citep{papineni-etal-2002-bleu, banerjee-lavie-2005-meteor,rei-etal-2020-comet, zhang2020bertscore} etc. The risk (training objective) for the system is:
\begin{equation}
    \begin{aligned}
    \Loss_{\text{MRT}} &= \sum_{s=1}^{S} \E_{\vec{y}|\vec{x}^{(s)};\theta} \left[ \Delta (\vec{y} , \vec{y}^{(s)}) \right] \\
            &= \sum_{s=1}^{S} \sum_{\vec{y}\in C(\vec{x})} P(\vec{y}| \vec{x}^{(s)};\theta) \Delta (\vec{y} , \vec{y}^{(s)}) \\
    \hat{\theta}_{\text{MRT}} &= \underset{\theta}{\text{argmin}}  \{\Loss_{\text{MRT}}(\theta)\} 
    \end{aligned}
\end{equation}

where $C(\vec{x}^{(s)})$ is the set of all possible candidate translation by the system. \citet{shen-etal-2016-minimum} shows that partial of risk $\Loss_{MRT}(\theta)$ with respect to a model parameter $\theta_i$ does not need to differentiate $\Delta (\vec{y} , \vec{y}^{(s)})$:
\begin{equation}
    \begin{split}
        \frac{\partial \Loss_{\text{MRT}}(\theta)}{\partial \theta_i} = \sum_{s=1}^{S} \E_{\vec{y}|\vec{x}^{(s)};\theta} \biggl[ \Delta (\vec{y} , \vec{y}^{(s)}) \times \\
        \sum_{n=1}^{N^{(s)}} \frac{\partial P(\vec{y}_{n}^{(s)} | \vec{x}^{(s)},  \vec{y}_{<n}^{(s)};\theta) / \partial \theta_i }
        {P(\vec{y}_{n}^{(s)} | \vec{x}^{(s)},  \vec{y}_{<n}^{(s)};\theta)} \biggr]
    \end{split}
\end{equation}

        
Hence MRT allows an arbitrary scoring function $\Delta$ to be used, whether it is differentiable or not. In our experiments, we use MRT with two metrics, BLEU \cite{papineni-etal-2002-bleu}---the standard in machine translation and COMET \cite{rei-etal-2020-comet}---a newly proposed neural-based evaluation metric that correlates better with human judgement. 

\paragraph{Adversarial Attack}
Adversarial attacks generate samples that closely match input while dramatically distorting the model output. The samples can be generated by either a white-box or a black-box model. Black-box methods do not have access to the model while white-box methods have such access. A set of adversarial samples are generated by:
\begin{equation}
    \{ \vec{x}' | \Reward(\vec{x}',\vec{x}) \leq \epsilon, \underset{\vec{x}'}{\text{argmax}} J(\vec{x}', \vec{y}; \theta) \}
\end{equation}
where $J(\cdot)$ is the probability of a sample being adversarial and $\Reward(\vec{x}', \vec{x})$ computes the degree of imperceptibility of perturbation $\vec{x}'$ compared to original input $\vec{x}$. The smaller the $\epsilon$, the less noticeable the perturbation is. In our system, $J(\cdot)$ not only focuses on attacking the forward model, but also uses the backward model to constrain the direction of gradient update and maintain source-side semantic similarity.

\section{Approach: Doubly Trained NMT for Adversarial Sample Generation}\label{section:approach}

\begin{figure*}[h]
    \centering
    \begin{tabular}{c}
    \includegraphics[width=1\textwidth]{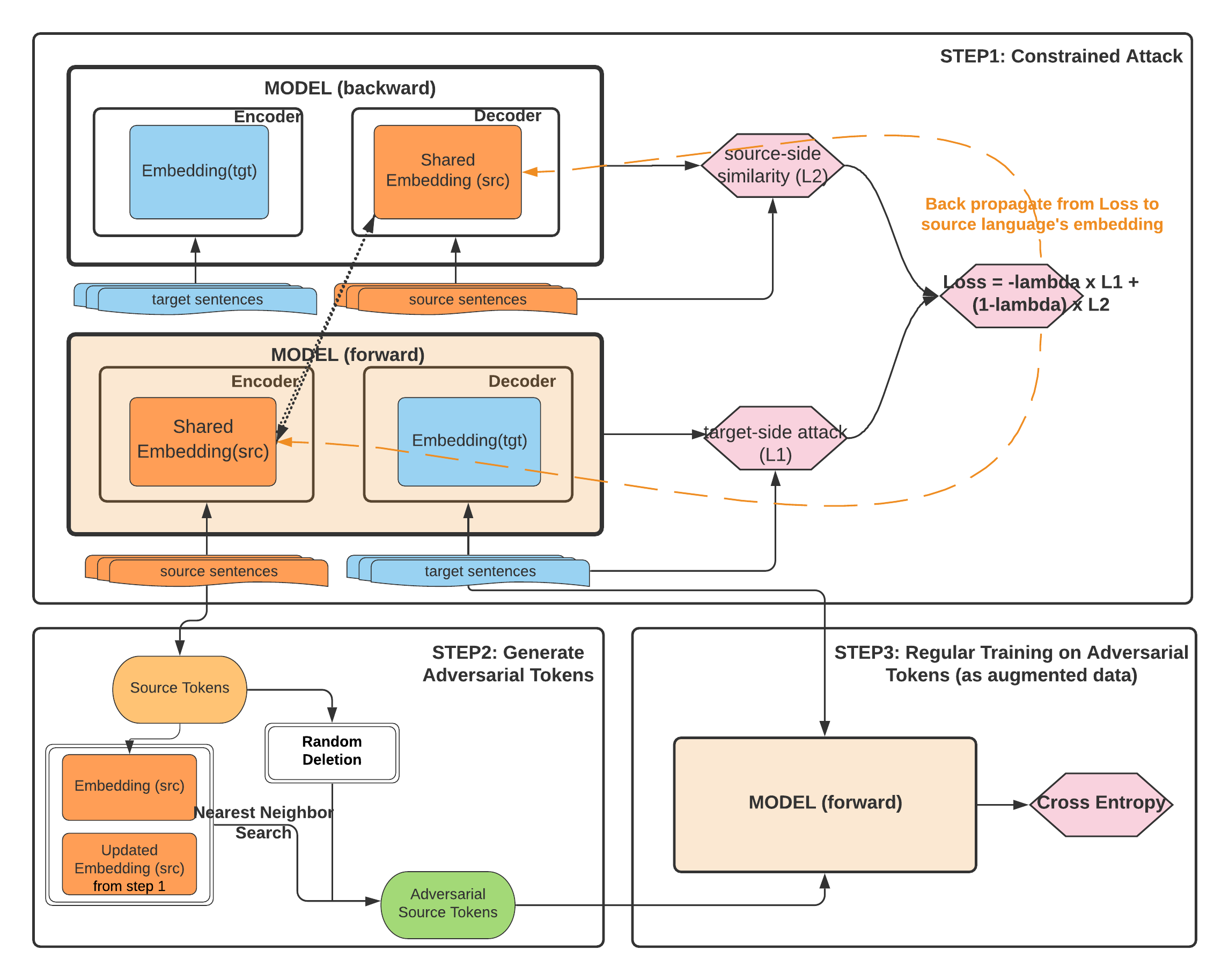} \\
    \end{tabular}
    \caption{Visual explanation of our adversarial augmentation algorithm. Step 1: Forward and backward models are trained simultaneously and attacked by the combined objective function. (The shared embedding is modified). Step 2: input source tokens are randomly deleted or replaced by nearest neighbor search to generate adversarial samples. Step 3: forward model is trained on adversarial samples.
    \label{fig:method}}
\end{figure*}

We aim to to generate adversarial samples that both preserve input's semantic meaning and decrease the performance of a NMT model.  We propose a doubly-trained system that involves two models of opposite translation direction (denote the forward model as $\theta_{st}$ and the backward model as $\theta_{ts}$). Our algorithm will train and update $\theta_{st}, \theta_{ts}$ simultaneously. Note that both models are pretrained before they are used for adversarial augmentation so that they can already produce good translations. Our algorithm has three steps as shown in \autoref{fig:method}.

\paragraph{Step 1 -- Constrained Attack to Update Embedding}
The first step is to attack the system and update the source embedding. We train the models with Negative Log-Likelihood (NLL) or MRT and combine the loss from two models as our final loss function to update the shared embedding. We denote the loss for $\theta_{st}$ as $\Loss_1$ and loss for $\theta_{ts}$ as $\Loss_2$. Because we want to attack the forward model and preserve translation quality for the backward model, we make our final loss $$\Loss = -\lambda \Loss_1 + (1-\lambda) \Loss_{2}$$ where $\lambda \in [0,1]$ and is used as the weight to decide whether we focus on punishing the forward model (large $\lambda$) or preserving the backward model (small $\lambda$). When we use NLL as training objective, we have $\Loss_1 = \textbf{NLL}(\vec{x}^{(s)}, \vec{y}^{(s)}, \theta_{st})$ where $\vec{x}^{(s)}$ is the input sentences, $\vec{y}^{(s)}$ is the gold standard translation and \textbf{NLL($\cdot$)} is the Negative Log-Likelihood function that computes a loss based on training data $\vec{x}^{(s)}, \vec{y}^{(s)}$ and model parameter $\theta_{st}$. Similarly we have $\Loss_2 =\textbf{NLL}(\vec{y}^{(s)}, \vec{x}^{(s)}, \theta_{ts}) $

We also experimented with MRT in our doubly-trained system because we wonder if taking sentence-level scoring functions like BLEU or COMET would help improve adversarial samples' quality. For model $\theta_{st}$, we feed in source sentences $\vec{x}^{(s)}$ and we infer a set of possible translation $S(\vec{x}^{(s)})$ as the subset of full sample space. The loss (risk) of our prediction is therefore calculated as:
\begin{equation}
    \begin{aligned}
    \Loss_1 &= \sum_{s=1}^{S} \E_{\vec{y}|\vec{x}^{(s)};\theta_{st}} \left[ \Delta (\vec{y} , \vec{y}^{(s)}) \right] \\
            &= \sum_{s=1}^{S} \sum_{\vec{y} \in S(\vec{x}^{(s)})} Q(\vec{y} | \vec{x}^{(s)};\theta_{st}, \alpha) \Delta (\vec{y} , \vec{y}^{(s)}) \\
    \end{aligned}
\end{equation}
where 
$$ Q(\vec{y} | \vec{x}^{(s)};\theta_{st}, \alpha) = \frac{P(\vec{y}|\vec{x}^{(s)}; \theta_{st})^\alpha }{\sum_{\vec{y}' \in S(\vec{x}^{(s)}}) P(\vec{y}'|\vec{x}^{(s)}; \theta_{st})^\alpha }$$

The value $\alpha$ here controls the sharpness of the formula and we follow \citet{shen-etal-2016-minimum}  to use $\alpha = 5e^{-3}$ throughout our experiments. To sample the subset of full inference space $S(\vec{x}^{(s)})$, we use Sampling Algorithm \citep{shen-etal-2016-minimum} to generate k translation candidates for each input sentence (During inference time, the model outputs a probabilistic distribution over the vocabulary for each token and we sample a token based on this distribution). It is denoted as $\textbf{Sample}(\vec{x}^{(s)},\theta, k)$ in our \hyperref[algo:embed-update]{Algorithm \ref*{algo:embed-update}}. Similarly, for model $\theta_{ts}$, we feed in the reference sentences of our parallel data and generate a set of possible translation $S(\vec{y}^{(s)})$ in source language. We compute the loss (risk) of source-side similarity as:
\begin{equation}
    \begin{aligned}
    \Loss_2 &= \sum_{s=1}^{S} \E_{\vec{x}|\vec{y}^{(s)};\theta_{ts}} \left[ \Delta (\vec{x} , \vec{x}^{(s)}) \right] \\
            &= \sum_{s=1}^{S} \sum_{\vec{x} \in S(\vec{y}^{(s)})} Q(\vec{x} | \vec{y}^{(s)};\theta_{ts}, \alpha) \Delta (\vec{x} , \vec{x}^{(s)}) \\
    \end{aligned}
\end{equation}

\noindent After computing loss using MRT or NLL, we have 
$$\Loss(\theta_{st}, \theta_{ts}) =  -\lambda \Loss_1 + (1-\lambda) \Loss_{2}$$ and we train the system to find 
$$\hat{\theta}_{st}, \hat{\theta}_{ts} = \underset{\theta_{st}, \theta_{ts}}{\text{argmin}} \{ \Loss (\theta_{st}, \theta_{ts})\} $$
To be updated from both risks, two models need to share some parameters since $\Loss_1$ only affects $\theta_{st}$ and $\Loss_2$ only updates $\theta_{ts}$. Because word embedding is a representation of input tokens, we make it such that the source-side embedding of $\theta_{st}$ and the target-side embedding of $\theta_{ts}$ is one shared embedding. We do so because they are both representations of source language in our translation and we can use it to generate adversarial tokens for source sentences in step 2. We also freeze all other layers in two models. Thus, when we update the model parameter $\theta_{st}, \theta_{ts}$, we only update the shared embedding of source language. The process described above is summarized in \hyperref[algo:embed-update]{Algorithm \ref*{algo:embed-update}}.

\paragraph{Step 2 -- Perturb input sentences to generate adversarial tokens}
After updating the shared embedding, we can use the updated embedding to generate adversarial tokens. We introduce two kinds of noise into input sentences to generate adversarial samples: random deletion and simple replacement. Random deletion is introduced by randomly deleting some tokens in the input sentences while simple replacement is introduced by using nearest neighbor search with original and updated embedding. Our implementation is described in \hyperref[section:adv-gen]{Section 4.3.1}. 


\paragraph{Step 3 -- Train on adversarial Samples}
After generating adversarial tokens from step 2, we directly train the forward model on them with the NLL loss function.
\begin{algorithm}
\caption{Update model embedding}
\begin{algorithmic} \label{algo:embed-update}
\STATE \textbf{Input:} Pretrained Models $\theta_{st}$ and $\theta_{ts}$, Max Number of Epochs E, Sample Size K, Sentence-Level Scoring Metric M
\STATE \textbf{Output:} Updated Models $\theta_{st}$ and $\theta_{ts}$ (only the shared embedding is updated)
\WHILE{$\theta_{st}$, $\theta_{ts}$ \text{not Converged} \textbf{and} $e \leq E$}
\FOR{$(\vec{x}^{(i)}, \vec{y}^{(i)}), 1 < i \leq S$}
\IF{using MRT as objective}

\STATE /* sample and compute the risk */
\STATE$S(\vec{x}^{(i)}) = \textbf{Sample}(\vec{x}^{(i)}, \theta_{st}, K)$ 
\STATE$\Loss_1 \gets \textbf{MRT}(S(\vec{x}^{(i)}), M, \vec{y}^{(i)}) $
\STATE /* Repeat for another direction */
\STATE$S(\vec{y}^{(i)}) = \textbf{Sample}(\vec{y}^{(i)}, \theta_{ts}, K)$ 
\STATE$\Loss_2 \gets \textbf{MRT}(S(\vec{y}^{(i)}), M, \vec{x}^{(i)}) $

\ELSIF{using NLL as objective}

\STATE $\Loss_1 \gets \textbf{NLL}(\vec{x}^{(i)}, \vec{y}^{(i)}, \theta_{st}) $
\STATE $\Loss_2 \gets \textbf{NLL}(\vec{x}^{(i)}, \vec{y}^{(i)}, \theta_{ts}) $

\ENDIF
\STATE $\Loss (\theta_{st}, \theta_{ts}) = -\lambda \Loss_1 + (1-\lambda) \Loss_2$
\STATE $\theta_{st}, \theta_{ts} \gets \nabla_{Emb} \Loss (\theta_{st}, \theta_{ts})$
\ENDFOR
\ENDWHILE
\end{algorithmic}
\end{algorithm}

\section{Experiment}
\subsection{Pretrained Model Setup}\label{section:pretrain}
We pretrain the standard Transformer \cite{vaswani2017attention} base model implemented in fairseq \cite{ott2019fairseq}. The hyper-parameters follow the \texttt{transformer-en-de} setup from fairseq and our script is shown in Appendix, \autoref{fig:pretrain_train}. We experimented on three different language pairs: Chinese-English (zh-en), German-English (de-en), and French-English (fr-en). For each language pair, two models are pretrained on the same training data using the same hyper-parameters and they share the embedding of source language. For example, for Chinese-English, we first train the forward model (zh-en) from scratch. Then we freeze the source language (zh)'s embedding from forward model and use it to pretrain our backward model (en-zh). The training data used for three languages pairs are: 
\begin{enumerate}
    \item zh-en: WMT17 \cite{bojar-etal-2017-findings} parallel corpus (except UN) for training, WMT2017 and 2018 \texttt{newstest} data for validation, and WMT2020 \texttt{newstest} for evaluation.
    \item de-en: WMT17 parallel corpus for training,  WMT2017 and 2018 \texttt{newstest} data for validation, and WMT2014 \texttt{newstest} for evaluation.
    \item fr-en: WMT14 \cite{bojar-EtAl:2014:W14-33} parallel corpus (except UN) for training, WMT2015 \texttt{newdicussdev} and \texttt{newsdiscusstest} for validation, and WMT2014 \texttt{newstest} for evaluation.
\end{enumerate}
For Chinese-English parallel corpus, we used a sentencepiece model of size 20k to perform BPE. For German-English and French-English data, we followed preprocessing scripts \footnote{github.com/pytorch/fairseq/tree/master/examples/translation} on fairseq and used subword-nmt of size 40k to perform BPE. We need two validation sets because in our experiment, we fine-tune the model with our adversarial augmentation algorithm on one of the validation set and use the other for model selection. After pretraining stage, the transformer models' performances on test sets are shown in \autoref{fig:pretrain-stats}. The evaluation of BLEU score is computed by SacreBLEU\footnote{Signature included in Appendix, \autoref{section:signature}} \cite{post-2018-call}.

\begin{table}
\centering
\begin{tabular}{llllll}
\toprule
\textbf{lang} & \textbf{BLEU} &\textbf{lang} & \textbf{BLEU} &\textbf{lang} & \textbf{BLEU}\\
\midrule
zh-en & 22.8 & de-en & 30.2 & fr-en & 34.5 \\
en-zh & 36.0 & en-de & 24.9 & en-fr & 35.3 \\
\bottomrule
\end{tabular}
\caption{\label{pretrain-model}
Pretrained baseline models' BLEU score
}
\label{fig:pretrain-stats}
\end{table}

\subsection{Doubly Trained System for Adversarial Attack}\label{section:adv-attack}
Our adversarial augmentation algorithm has three steps: the first step is performing a constrained adversarial attack while the remaining steps generate and train models on augmentation data. In this section, we experiment with only the first step and test if \hyperref[algo:embed-update]{Algorithm \ref*{algo:embed-update}} can generate meaning-preserving update on the embedding. Our objective function $\Loss (\theta_{st}, \theta_{ts}) = -\lambda \Loss_1 + (1-\lambda) \Loss_2$ is a combination of two rewards from forward and backward models. The expectation is that after the perturbation on the embedding, the forward model's performance would drastically decrease (because it's attacked) and the backward model should still translate reasonably well (because the objective function preserves the source-side semantic meaning). We perform the experiment on Chinese-English and results are shown in \autoref{fig:adv-data}. We find that models corroborate to our expectation: After 15 epochs, the forward (zh-en) model's performance drops significantly while the backward (en-zh) model's performance barely decreases. After 20 epochs, the forward model is producing garbage translation while the backward model is still performing well. 

 
\begin{table}
\centering
\begin{tabular}{lllll}
\toprule
\textbf{\#Epochs} & \textbf{BLEU (zh-en)} & \textbf{BLEU (en-zh)} \\
\midrule
10 & 20.1  & 34.0 \\
15 & 10.9  & 32.4 \\
20 & 0.3 &  33.5 \\
30 & 0.0 &  32.1 \\
\bottomrule
\end{tabular}
\caption{
Forward and backward models' performance (of Chinese and English) after adversarial attack using MRT as training objective, described in \hyperref[algo:embed-update]{Algorithm \ref*{algo:embed-update}}.}
\label{fig:adv-data}
\end{table}


\subsection{Doubly Trained System for Data Augmentation}
From \hyperref[section:adv-attack]{Section 4.2}, we have verified that the first step of our adversarial augmentation training is effective to generate meaning-preserving perturbation on the word embedding. We then perform all three steps of our algorithm to investigate whether it is effective as an augmentation technique, which is the focus of this work. In the following subsections, we show, in detail, how the adversarial samples and noisy test data are generated, as well as the test results of our models.

\begin{table*}
\centering
\small\begin{tabular}{lllllllllll}
\toprule

\textbf{Model (ZH-EN)} & \textbf{RD10}  & \textbf{RD15} & \textbf{RD20} & \textbf{RD25} & \textbf{RD30} & \textbf{RP10} & \textbf{RP15} & \textbf{RP20} & \textbf{RP25}  & \textbf{RP30}\\
\midrule
Baseline &25$\%$&36$\%$&46$\%$&55$\%$&63$\%$&8$\%$&14$\%$&19$\%$&22$\%$&25$\%$\\
Finetune &23$\%$&33$\%$	&42$\%$&52$\%$	&60$\%$	&8$\%$&11$\%$&14$\%$&17$\%$&21$\%$\\
Simple Replacement &23$\%$&33$\%$&	41$\%$&	51$\%$&	59$\%$&6$\%$&8$\%$&10$\%$&12$\%$&15$\%$\\
Dual NLL&\textbf{21}$\%$&\textbf{31}$\%$&	\textbf{40}$\%$&	\textbf{49}$\%$&\textbf{56}$\%$&4$\%$&6$\%$&8$\%$&\textbf{10}$\%$&\textbf{12}$\%$\\
Dual BLEU& 23$\%$&33$\%$&42$\%$&51$\%$&58$\%$&4$\%$&6$\%$&9$\%$&11$\%$&13$\%$\\
Dual COMET&22$\%$&32$\%$&41$\%$&50$\%$&58$\%$&\textbf{4}$\%$&\textbf{6}$\%$&\textbf{8}$\%$&10$\%$&13$\%$\\
\bottomrule
\end{tabular}

\small\begin{tabular}{lllllllllll}
\toprule
\textbf{Model (DE-EN)}& \textbf{RD10}  & \textbf{RD15} & \textbf{RD20} & \textbf{RD25} & \textbf{RD30} & \textbf{RP10} & \textbf{RP15} & \textbf{RP20} & \textbf{RP25}  & \textbf{RP30}\\
\midrule
Baseline &43$\%$&51$\%$&60$\%$&68$\%$&74$\%$&31$\%$&34$\%$&37$\%$&40$\%$&44$\%$\\
Finetune &42$\%$&50$\%$	&58$\%$&67$\%$	&73$\%$	&31$\%$&34$\%$&37$\%$&40$\%$&44$\%$\\
Simple Replacement &42$\%$&50$\%$&	59$\%$&	66$\%$&	72$\%$&30$\%$&32$\%$&35$\%$&37$\%$&40$\%$\\
Dual NLL&42$\%$&49$\%$&\textbf{56}$\%$&	\textbf{63}$\%$&\textbf{69}$\%$&29$\%$&31$\%$&33$\%$&35$\%$&37$\%$\\
Dual BLEU& \textbf{41}$\%$&49$\%$&57$\%$&64$\%$&71$\%$&\textbf{28}$\%$&\textbf{30}$\%$&\textbf{33}$\%$&\textbf{35}$\%$&\textbf{37}$\%$\\
Dual COMET&42$\%$&\textbf{48}$\%$&57$\%$&64$\%$&70$\%$&29$\%$&31$\%$&33$\%$&35$\%$&38$\%$ \\
\bottomrule
\end{tabular}

\small\begin{tabular}{lllllllllll}
\toprule
\textbf{Model (FR-EN)} & \textbf{RD10}  & \textbf{RD15} & \textbf{RD20} & \textbf{RD25} & \textbf{RD30} & \textbf{RP10} & \textbf{RP15} & \textbf{RP20} & \textbf{RP25}  & \textbf{RP30}\\
\midrule
Baseline &47$\%$&54$\%$&61$\%$&67$\%$&74$\%$&38$\%$&40$\%$&44$\%$&47$\%$&50$\%$\\
Finetune &47$\%$&54$\%$	&60$\%$&67$\%$	&73$\%$	&37$\%$&40$\%$&44$\%$&48$\%$&49$\%$\\
Simple Replacement &45$\%$&53$\%$&	60$\%$&	66$\%$&	73$\%$&35$\%$&37$\%$&40$\%$&43$\%$&46$\%$\\
Dual NLL&\textbf{45}$\%$&\textbf{52}$\%$&	59$\%$&	65$\%$&71$\%$&35$\%$&37$\%$&40$\%$&43$\%$&45$\%$\\
Dual BLEU& 45$\%$&52$\%$&59$\%$&66$\%$&72$\%$&35$\%$&\textbf{36}$\%$&\textbf{39}$\%$&\textbf{41}$\%$&\textbf{44}$\%$\\
Dual COMET&45$\%$&52$\%$&\textbf{58}$\%$&\textbf{65}$\%$&\textbf{71}$\%$&\textbf{34}$\%$&37$\%$&39$\%$&42$\%$&44$\%$\\
\bottomrule
\end{tabular}
\caption{\label{adversarial-as-data-augmentation-bleu}
Models' performance on noisy synthetic data generated from random deletion (RD) and simple replacement (RP). Number after RD/RP is the percentage of noise introduced in clean data (e.g RD15 is the test set generated by randomly deleting 15$\%$ of clean test data). Generated translation are measured by $\Delta \text{BLEU}$. We define $\text{BLEU}(x)$ as the BLEU score evaluated on test dataset x (e.g. RD10), $\Delta \text{BLEU}(x) = 1 - \frac{\text{BLEU}(x)}{\text{BLEU}(\text{clean})}, $where $\text{BLEU(clean)}$ is BLEU score of the model evaluated on the clean dataset. The higher the $\Delta$BLEU, the worse the model on noisy data. The details of the six models and analysis are included in \hyperref[section:analysis]{Section 4.3.3}.}
\end{table*}

\subsubsection{Adversarial Sample Generation:}\label{section:adv-gen}
To generate adversarial tokens (due to the discrete nature of natural languages), we resort to cosine similarity. Let model embedding be $E$ before the embedding update, and $E'$ after the update from \hyperref[algo:embed-update]{Algorithm \ref*{algo:embed-update}}. Let the vocab be $V$ and let input sentence be $S = \{s_1, s_2, \cdots s_n\}$ For each token $s_i \in S, s_i \notin \{EOS, BOS, PAD\}$, three actions are possible:
\begin{enumerate}
    \item no perturbation, with probability $P_{np}$
    \item perturbed into most similar token by updated embedding with probability $P_{rp}$
    \item perturbed to be empty token (deleted at this position) with probability $P_{rd} = 1 - P_{rp}$  
\end{enumerate}
Throughout our experiments, we set the hyper-parameters as $P_{np}=0.7, P_{rp}= 0.8, P_{rd}= 0.2$. That means each token has 30 percent chance to be perturbed, and if that's the case, it has 80 percent chance to be replaced by a similar token and 20 percent chance to be deleted. For no-perturbation or deletion case, it's straightforward to implement. For replacement, we compute $s_{i}'$ (the adversarial token of $s_i$) by cosine similarity: $s_{i}' = \underset{v\in V, v \neq s_i}{argmax}(\frac{E'[s_i]}{|E'[s_i]|} \cdot \frac{E[v]}{|E[v]|} )$. 

For the credibility of this hyper-parameter setup, we perform a grid search over 9 possible combinations: $\underset{P_{np}}{(0.6, 0.7, 0.8)} \times \underset{P_{rp}}{(0.6, 0.7, 0.8)}$ and found that the difference in performance is mostly due to model type instead of probability setup. Details of grid search can be found in Appendix (\autoref{fig:grid-prob}).

\subsubsection{Noisy Test Data}\label{section:noisy-test}
In order to test the model robustness after training on augmented data, we create synthetic noise data from test data of different languages mentioned in \hyperref[section:pretrain]{Section 4.1}. We follow the practice from \citet{niu-etal-2020-evaluating} and perturb the test data to varying degree, ranging from 10$\%$ to 30$\%$. We focus on two kinds of noise: random deletion and simple replacement. The procedure we introduce synthetic noise into clean test data is the same as the procedure described in \hyperref[section:adv-gen]{Section 4.3.1}. The only difference is in the case of simple replacement: instead of having original embedding $E$ and updated embedding $E'$ from attack step, we only have the embedding $E$ from pretrained model and there is no training step to update it. The perturbed token is therefore computed by $s_{i}' = \underset{v\in V, v\neq s_i}{argmax}(\frac{E[s_i]}{|E[s_i]|} \cdot \frac{E[v]}{|E[v]|} )$. Note that all operations are done in the subword level because that is how we preprocessed the data for Transformer model. 

\begin{table*}
\centering
\small\begin{tabular}{lllllllllll}
\toprule
\textbf{Model (ZH-EN)} & \textbf{RD10}  & \textbf{RD15} & \textbf{RD20} & \textbf{RD25} & \textbf{RD30} & \textbf{RP10} & \textbf{RP15} & \textbf{RP20} & \textbf{RP25}  & \textbf{RP30}\\
\midrule
Baseline &99$\%$&158$\%$&210$\%$&278$\%$&342$\%$&48$\%$&68$\%$&95$\%$&116$\%$&137$\%$\\
Finetune &66$\%$&105$\%$	&143$\%$&189$\%$	&236$\%$	&30$\%$&41$\%$&56$\%$&67$\%$&80$\%$\\
Simple Replacement &63$\%$&105$\%$&	139$\%$&	184$\%$&	230$\%$&19$\%$&27$\%$&36$\%$&48$\%$&62$\%$\\
Dual NLL&64$\%$&103$\%$&	\textbf{135}$\%$&	\textbf{176}$\%$&225$\%$&20$\%$&29$\%$&37$\%$&47$\%$&56$\%$\\
Dual BLEU& 64$\%$&102$\%$&138$\%$&181$\%$&\textbf{224}$\%$&\textbf{18}$\%$&\textbf{26}$\%$&\textbf{35}$\%$&\textbf{46}$\%$&\textbf{56}$\%$\\
Dual COMET&\textbf{63}$\%$&\textbf{102}$\%$&136$\%$&180$\%$&227$\%$&18$\%$&27$\%$&38$\%$&47$\%$&57$\%$ \\
\bottomrule
\end{tabular}

\small\begin{tabular}{lllllllllll}
\toprule
\textbf{Model (DE-EN)}& \textbf{RD10}  & \textbf{RD15} & \textbf{RD20} & \textbf{RD25} & \textbf{RD30} & \textbf{RP10} & \textbf{RP15} & \textbf{RP20} & \textbf{RP25}  & \textbf{RP30}\\
\midrule
Baseline &124$\%$&159$\%$&196$\%$&230$\%$&265$\%$&76$\%$&88$\%$&99$\%$&113$\%$&127$\%$\\
Finetune &116$\%$&150$\%$	&186$\%$&220$\%$	&255$\%$	&72$\%$&83$\%$&95$\%$&108$\%$&122$\%$\\
Simple Replacement &113$\%$&145$\%$&	179$\%$&	212$\%$&	245$\%$&\textbf{68}$\%$&78$\%$&88$\%$&98$\%$&109$\%$\\
Dual NLL&114$\%$&146$\%$&	177$\%$&208$\%$&241$\%$&71$\%$&80$\%$&88$\%$&97$\%$&108$\%$\\
Dual BLEU& \textbf{113}$\%$&\textbf{144}$\%$&\textbf{176}$\%$&\textbf{208}$\%$&\textbf{240}$\%$&69$\%$&\textbf{78}$\%$&\textbf{86}$\%$&\textbf{95}$\%$&\textbf{106}$\%$\\
Dual COMET&114$\%$&144$\%$&177$\%$&209$\%$&242$\%$&70$\%$&79$\%$&87$\%$&96$\%$&107$\%$ \\
\bottomrule
\end{tabular}

\small\begin{tabular}{lllllllllll}
\toprule
\textbf{Model (FR-EN)} & \textbf{RD10}  & \textbf{RD15} & \textbf{RD20} & \textbf{RD25} & \textbf{RD30} & \textbf{RP10} & \textbf{RP15} & \textbf{RP20} & \textbf{RP25}  & \textbf{RP30}\\
\midrule
Baseline &132$\%$&156$\%$&178$\%$&204$\%$&228$\%$&104$\%$&113$\%$&122$\%$&132$\%$&142$\%$\\
Finetune &122$\%$&147$\%$	&171$\%$&197$\%$	&221$\%$	&91$\%$&100$\%$&109$\%$&119$\%$&128$\%$\\
Simple Replacement &121$\%$&144$\%$&	167$\%$&	193$\%$&	217$\%$&89$\%$&96$\%$&104$\%$&113$\%$&120$\%$\\
Dual NLL&120$\%$&143$\%$&	165$\%$&	\textbf{190}$\%$&\textbf{213}$\%$&89$\%$&97$\%$&105$\%$&112$\%$&121$\%$\\
Dual BLEU& 121$\%$&144$\%$&167$\%$&192$\%$&216$\%$&89$\%$&96$\%$&104$\%$&110$\%$&\textbf{117}$\%$\\
Dual COMET&\textbf{120}$\%$&\textbf{143}$\%$&\textbf{165}$\%$&191$\%$&214$\%$&\textbf{88}$\%$&\textbf{95}$\%$&\textbf{102}$\%$&\textbf{110}$\%$&118$\%$\\
\bottomrule
\end{tabular}
\caption{\label{adversarial-as-data-augmentation-comet}
Models' performance on noisy synthetic data generated from random deletion (RD) and simple replacement (RP). Set-up is the same as \autoref{adversarial-as-data-augmentation-bleu} except that evaluation metric is COMET instead of BLEU, so we show $\Delta$COMET here. Note that $\Delta$COMET can go over 100\% because COMET score can be negative.}
\end{table*}

\subsubsection{Result Analysis}\label{section:analysis}
We show our results in \autoref{adversarial-as-data-augmentation-bleu} and \autoref{adversarial-as-data-augmentation-comet}. For each language pair, there are 6 types of models in each plot:
\begin{enumerate}
    \item \textbf{baseline model}: pretrained forward (src-tgt) model
    \item \textbf{fine-tuned model}: baseline model fine-tuned on validation set using NLL loss
    \item \textbf{simple replacement model}: baseline model fine-tuned on adversarial tokens. This model is fine-tuned using procedure described in \autoref{fig:method} without the first step. Adversarial samples are generated the same way we introduce noise into clean test data \hyperref[section:noisy-test]{(Section 4.3.2)} because there is no updated embedding from attack ($s_{i}' = \underset{v\in V, v\neq s_i}{argmax}(\frac{E[s_i]}{|E[s_i]|} \cdot \frac{E[v]}{|E[v]|} )$).
    \item \textbf{dual-nll model}: baseline model fine-tuned on adversarial tokens generated by doubly-trained system with NLL as training objective. 
    \item \textbf{dual-bleu model}: baseline model fine-tuned on adversarial tokens generated by doubly-trained system with MRT as training objective. It uses BLEU as the metric to compute MRT risk. 
    \item \textbf{dual-comet model}: same as dual-bleu model above except that it uses COMET as the metric for MRT risk.
\end{enumerate}
\noindent We show the percentage of change evaluated by BLEU and COMET on \autoref{adversarial-as-data-augmentation-bleu} and \autoref{adversarial-as-data-augmentation-comet}, computed by
$$\Delta \text{Metric}(x) = 1 - \frac{\text{Metric}(x)}{\text{Metric}(\text{clean})}$$
where the metric can be BLEU or COMET, and x represents the test data used, as explained in \autoref{adversarial-as-data-augmentation-bleu}. As the ratio of noise increases, $\text{Metric}(x)$ decreases, which increases $\Delta \text{Metric}(x)$. Therefore, robust models resist to the increase of noise ratio and have lower  $\Delta \text{Metric}(x)$. From both tables, we find that doubly-trained models (dual-nll, dual-bleu, and dual-comet) are more robust than the other models regardless of test data, evaluation metrics, or language pairs used.

For any NMT model tested on the same task evaluated by two metrics (any corresponding row in \autoref{adversarial-as-data-augmentation-bleu} and \autoref{adversarial-as-data-augmentation-comet}), BLEU and COMET give similar results though COMET have a larger difference among models because its percentage change is more drastic. We performed tests using COMET in addition to BLEU because we use MRT with BLEU and COMET in attack step and we want to see if performances of dual-comet and dual-bleu model differ under either evaluation metric. From our results, there is no noticeable difference. This might happen because we used a small learning rate for embedding update in attack step or simply because BLEU and COMET give similar evaluation. 

Comparing the results in \autoref{adversarial-as-data-augmentation-bleu} and \autoref{adversarial-as-data-augmentation-comet}, we see margins of models' performance are bigger when evaluated on noisy test data generated with replacement (RP columns). This is expected because random deletion introduces more noise than replacement and it's hard for models to defend against it. Therefore, though doubly trained systems have substantial improvement against other models when noise type is simple replacement, they gain less noticeable advantage in random deletion test.

Lastly, when we compare across doubly-trained systems (dual-nll, dual-bleu, and dual-comet), we see that they are comparable to each other within a margin of 3 percent. This implies that incorporating sentence-level scoring metric with MRT might not be as effective as we thought for word-level adversarial augmentation. Considering the training overhead of MRT, NLL-based doubly-trained system will be more effective to enhance model robustness against synthetic noise. 


\section{Related Work}
Natural and synthetic noise affect models in translation \cite{belinkov2018synthetic} and adversarial perturbation is commonly used to evaluate and improve model robustness in such cases. Various adversarial methods are researched for robustness, some use adversarial samples as regularization \cite{sato-etal-2019-effective}, some incorporate it with reinforcement learning \cite{zou2020reinforced}, and some use it for data augmentation. When used for augmentation, black-box adversarial methods tend to augment data by introducing noise into training data. For most of the time, simple operations such as random deletion/replacement/insertion are used for black-box attack \cite{karpukhin2019training}, though such operations can be used as white-box attack with gradients as well \cite{ebrahimi2018adversarial}. 

Most white-box adversarial methods use different architecture to attack and update model \cite{michel2019evaluation,cheng2020advaug,cheng2019robust}, and from which, generate augmented data. White-box adversarial methods gives more flexible modification for the token but at the same time become time consuming, making it infeasible for some cases when speed matters. Though it is commonly believed that white-box adversarial methods have higher capacity, there is study that shows simple replacement can be used as an effective and fast alternative to white-box methods where it achieves comparable (or even better) results for some synthetic noise \cite{takase2021rethinking}. This finding correlates with our research to some degree because we also find replacement useful to improve model robustness, though we perform replacement by most similar token instead of sampling a random token.

Compared to synthetic noise, natural noise is harder to test and defend against because natural noise can take various forms including grammatical errors, informal languages and orthographic variations among others. Some datasets such as MTNT \cite{michel2018mtnt} are proposed to be used as a testbed for natural noise. In recent WMT Shared Task on Robustness \cite{li2019findings}, systems and algorithms have been proposed to defend against natural noise, including domain-sensitive data mixing \cite{zheng-etal-2019-robust}, synthetic noise introduction \cite{karpukhin2019training, grozea-2019-system} and placeholder mechanism \cite{murakami-etal-2019-ntts} for non-standard tokens like emojis and emoticons. In this paper, we focus on synthetic noise and will leave natural noise for future work.

\section{Conclusion}
In this paper, we proposed a white-box adversarial augmentation algorithm to improve model robustness. We use doubly-trained system to perform constrained attack and then train the model on adversarial samples generated with random deletion and gradient-based replacement. Experiments across different languages and evaluation metrics have shown consistent improvement for model robustness. In the future, we will investigate on our algorithm's effectiveness for natural noise and try to incorporate faster training objective such as Contrastive-Margin Loss \cite{shu2021reward}.

\bibliography{anthology,custom}
\bibliographystyle{acl_natbib}

\clearpage
\appendix
\section{Appendices}
\subsection{Pretrained model}
Hyper-parameter for Pretraining the transformers (same for three language pairs) is shown in \autoref{fig:pretrain_train}. Note that for the fine-tune model, we use the same hyper-parameter as in pretraining, and we simply change the data directory into validation set to tune the pretrained model.
\begin{figure*}[h]
    \begin{verbatim}
fairseq-train $DATADIR \
    --source-lang src \
    --target-lang tgt \
    --save-dir $SAVEDIR \
    --share-decoder-input-output-embed \
    --arch transformer_wmt_en_de \
    --optimizer adam --adam-betas '(0.9, 0.98)' --clip-norm 0.0 \
    --lr-scheduler inverse_sqrt \ 
    --warmup-init-lr 1e-07 --warmup-updates 4000 \
    --lr 0.0005 --min-lr 1e-09 \
    --dropout 0.3 --weight-decay 0.0001 \
    --criterion label_smoothed_cross_entropy --label-smoothing 0.1 \
    --max-tokens 2048 --update-freq 16 \
    --seed 2 \
    \end{verbatim}
    \caption{This setup is used for all pretrained models, regardless of the language pair}
    \label{fig:pretrain_train}
\end{figure*}

\subsection{Adversarial Attack on Chinese-English Model}
Adversarail Attacks are performed with hyper-parameters shown in \autoref{fig:adv_attack}
\begin{figure*}[h]
    \begin{verbatim}
fairseq-train $DATADIR \
    --source-lang src \
    --target-lang tgt \
    --save-dir $SAVEDIR \
    --share-decoder-input-output-embed \
    --train-subset valid \
    --arch transformer_wmt_en_de \
    --optimizer adam --adam-betas '(0.9, 0.98)' --clip-norm 0.0 \
    --lr-scheduler inverse_sqrt \ 
    --warmup-init-lr 1e-07 --warmup-updates 4000 \
    --lr 0.0005 --min-lr 1e-09 \
    --dropout 0.3 --weight-decay 0.0001 \
    --criterion dual_bleu --mrt-k 16 \
    --batch-size 2 --update-freq 64 \
    --seed 2 \
    --restore-file $PREETRAIN_MODEL \
    --reset-optimizer \
    --reset-dataloader \
    \end{verbatim}
    \caption{Note that criterion is called "dual bleu" and this is our customized criterion based on fairseq. It implements the doubly trained adversarial attack algorithm discussed in this paper with sample size 16 (mrt-k = 16).}
    \label{fig:adv_attack}
\end{figure*}

\subsection{Data Augmentation}
Hyper-parameter for fine-tuning the base model with proposed doubly-trained algorithm on validation set is shown in \autoref{fig:data_aug}

Note that the criterion is either "dual mrt" (using BLEU as metric for MRT), "dual comet" (using COMET as metric for MRT) or "dual nll" (using NLL as training objective). These are customized criterion that we wrote to implement our algorithm. 

\begin{figure*}[h]
    \begin{verbatim}
fairseq-train $DATADIR \
    -s $src -t $tgt \
    --train-subset valid \
    --valid-subset valid1 \
    --left-pad-source False \
    --share-decoder-input-output-embed \
    --encoder-embed-dim 512 \
    --arch transformer_wmt_en_de \
    --dual-training \
    --auxillary-model-path $AUX_MODEL \
    --auxillary-model-save-dir $AUX_MODEL_SAVE \
    --optimizer adam --adam-betas '(0.9, 0.98)' --clip-norm 0.0 \
    --lr-scheduler inverse_sqrt \
    --warmup-init-lr 0.000001 --warmup-updates 1000 \
    --lr 0.00001 --min-lr 1e-09 \
    --dropout 0.3 --weight-decay 0.0001 \
    --criterion dual_comet/dual_mrt/dual_nll --mrt-k 8 \
    --comet-route $COMET_PATH \
    --batch-size 4 \
    --skip-invalid-size-inputs-valid-test \
    --update-freq 1 \
    --on-the-fly-train --adv-percent 30 \
    --seed 2 \
    --restore-file $PRETRAIN_MODEL \
    --reset-optimizer \
    --reset-dataloader \
    --save-dir $CHECKPOINT_FOLDER \
    \end{verbatim}
    \caption{Script for using doubly trained system for data augmentation}
    \label{fig:data_aug}
\end{figure*}
BLEU score for doubly-trained model's performance on noisy test data is shown in \autoref{adversarial-as-data-augmentation-bleu} and COMET score is shown in \autoref{adversarial-as-data-augmentation-comet}. Note that sometimes the $\Delta$COMET can be larger than 100$\%$ because COMET score can go from positive to negative.

\subsection{Choosing Hyper-parameter: Grid Search}

\subsubsection{Grid Search for $\lambda$}
lambda is the hyper-parameter used to balance the weight for the two risks in our doubly trained system. Recall the formula of our objective function: $\Loss (\theta_{st}, \theta_{ts}) = \lambda \Reward_1 - (1-\lambda) \Reward_2$. We perform grid search over $\underset{\lambda}{(0.2, 0.5, 0.8)}$ using dual-bleu and dual-comet model. It can be shown in \autoref{fig:lamda-test-bleu} and \autoref{fig:lamda-test-comet} that $\lambda$ value does not have a large impact on evaluation results and we pick $\lambda = 0.8$ throughout the experiments.

\begin{table*}
    \centering
    \small
    \begin{tabular}{llll}
    \toprule
    $\lambda$ & BLEU(zh-en) & BLEU(de-en) & BLEU(fr-en) \\
    \midrule
    0.2 & 28.6 & 46.9 & 40.0\\
    0.5 & 28.5 & 47.1 & 39.9\\
    0.8 & 28.4 & 47.0 & 39.8\\
    \bottomrule
    \end{tabular}
    \caption{dual-bleu model's performance on varying $\lambda$ values}
    \label{fig:lamda-test-bleu}
\end{table*}

\begin{table*}
    \centering
    \small
    \begin{tabular}{llll}
    \toprule
    $\lambda$ & BLEU(zh-en) & BLEU(de-en) & BLEU(fr-en) \\
    \midrule
    0.2 & 28.6 & 47.1 & 39.8\\
    0.5 & 28.7 & 46.9 & 39.9\\
    0.8 & 28.5 & 46.8 & 39.8\\
    \bottomrule
    \end{tabular}
    \caption{dual-comet model's performance on varying $\lambda$ values}
    \label{fig:lamda-test-comet}
\end{table*}

\subsubsection{Grid Search for $P_{np}, P_{rp}$}
We perform grid search for $P_{np}$, the probability of not perturbing a token, and $P_{rp}$, the probability of replacing the token if decided to modify it. Our search space is $\underset{P_{np}}{(0.6, 0.7, 0.8)} \times \underset{P_{rp}}{(0.6, 0.7, 0.8)}$ and the results are shown in \autoref{fig:grid-prob}. Since there is no noticeable difference across various $P_{np}, P_{rp}$ values, we pick $P_{np}=0.7, P_{rp}=0.8$ throughout our experiments.

\begin{table*}
    \centering
    \small
    \begin{tabular}{lllll}
        \toprule
        model (zh-en) & & $P_{rp}=60$ & $P_{rp}=70$ & $P_{rp}=80$ \\
        \midrule
        simple replacement & $P_{np}=60$ & 26.8 &26.8 &26.8 \\
        & $P_{np}=70$ & 26.8 &26.9	&26.8 \\
        & $P_{np}=80$ &27.0	&27.1&	27.0 \\
        \midrule
        dual-bleu & $P_{np}=60$ & 28.1 &28.2 &28.2 \\
        & $P_{np}=70$ & 28.4 &28.4	&28.4 \\
        & $P_{np}=80$ &28.4	&28.5&	28.6 \\
        \midrule
        dual-comet & $P_{np}=60$ & 28.4 &28.5 &28.4 \\
        & $P_{np}=70$ & 28.4 &28.4	&28.4 \\
        & $P_{np}=80$ &28.6	&28.7&	28.7 \\
        \bottomrule
    
    \end{tabular}

    \begin{tabular}{lllll}
        \toprule
        model (de-en) &  & $P_{rp}=60$ & $P_{rp}=70$ & $P_{rp}=80$ \\
        \midrule
        simple replacement & $P_{np}=60$ & 43.8&	43.9&	43.9 \\
        & $P_{np}=70$ &44.0	&44.0	&44.0 \\
        & $P_{np}=80$ &44.3&	44.3&	44.3 \\
        \midrule
        dual-bleu & $P_{np}=60$&46.4&	46.6&	46.5 \\
        & $P_{np}=70$ & 46.7&	46.7&	47.0 \\
        & $P_{np}=80$ &47.2&	47.1&	47.3 \\
        \midrule
        dual-comet & $P_{np}=60$ &46.5&	46.6&	46.7 \\
        & $P_{np}=70$ & 46.7&	46.7&	46.8 \\
        & $P_{np}=80$ &47.2&	47.3	&47.3 \\
        \bottomrule
    \end{tabular}

    \begin{tabular}{lllll}
        \toprule
        model (fr-en) & & $P_{rp}=60$ & $P_{rp}=70$ & $P_{rp}=80$ \\
        \midrule
        simple replacement & $P_{np}=60$ & 37.6 &37.6 &37.6 \\
        & $P_{np}=70$ & 37.8 &37.7	&37.6 \\
        & $P_{np}=80$ &37.8	&37.8&37.7 \\
        \midrule
        dual-bleu & $P_{np}=60$ & 39.5&39.8 &39.6 \\
        & $P_{np}=70$ & 39.6&39.9	&39.9 \\
        & $P_{np}=80$ & 40.0&40.1 &40.1 \\
        \midrule
        dual-comet & $P_{np}=60$ & 39.9&39.7 &39.8 \\
        & $P_{np}=70$ & 39.9&39.7	&39.7 \\
        & $P_{np}=80$ & 40.0&40.1 &40.0 \\
        \bottomrule
    \end{tabular}
    \caption{Evaluation performance based on varying probability of modification and replacement. \\ $P_{rp}:\text{Probability of replacing the token}$, $P_{np}:\text{Probability of not perturbing a token.}$ $P_{np}=60$ means we only perturb 40 percent of the input tokens}
    \label{fig:grid-prob}
\end{table*}    

\subsection{SacreBleu Signature:}\label{section:signature}
The signature generated by SacreBleu is \textit{"nrefs:1|case:mixed|tok:13a|smooth:exp|version:1.5.1"}. When evaluated with Chinese test data, we manually tokenize the predictions from our en-zh model with \textbf{tok=sacrebleu.tokenizers.TokenizerZh()} before computing corpus bleu with SacreBleu. The implementation can be found in our code.\footnote{https://github.com/steventan0110/NMTModelAttack}

\end{document}